# Universal Empathy and Ethical Bias for Artificial General Intelligence

Alexey Potapov[1,2], Sergey Rodionov[1,3]

*[1]AIDEUS, Russia*

*[2]National Research University of Information Technology, Mechanics and Optics, St. Petersburg, Russia*

*[3]Aix Marseille Université, CNRS, LAM (Laboratoire d'Astrophysique de Marseille) UMR 7326, 13388, Marseille, France*

{potapov,rodionov}@aideus.com

# Universal Empathy and Ethical Bias for Artificial General Intelligence

Rational agents are usually built to maximize rewards. However, AGI agents can find undesirable ways of maximizing any prior reward function. Therefore value learning is crucial for safe AGI. We assume that generalized states of the world are valuable – not rewards themselves, and propose an extension of AIXI, in which rewards are used only to bootstrap hierarchical value learning. The modified AIXI agent is considered in the multi-agent environment, where other agents can be either humans or other "mature" agents, which values should be revealed and adopted by the "infant" AGI agent. General framework for designing such empathic agent with ethical bias is proposed also as an extension of the universal intelligence model. Moreover, we perform experiments in the simple Markov environment, which demonstrate feasibility of our approach to value learning in safe AGI.

**Introduction**

Intelligent agents should have some motivation or pursue some goals. Most works on AI assume that these goals are correctly stated, and one can focus on problem solving. However, the problem of motivation is much more urgent in the case of AGI agents. Indeed, it is almost impossible to set such "basic" prior goal as survival. It is much easier to use somatic pain and pleasure for motivation, but this motivation will not guarantee optimal survivability. This problem is even more appreciable in the context of safe AGI, within which motivation and goal issues and their desirable realization are of first priority.

Different approaches to safe AGI have been already proposed. Some excellent surveys on this topic exist, e.g. (Sotala et al., 2013), and there is no need to repeat them, but it should be concluded that different approaches aimed at complete solution of the safety problem can be expressed in terms of value functions.

Value functions don't solve the problem, but help to state it. Indeed, the problem of complex values still remains (Yudkowsky, 2011). Safe value functions should be expressed in terms of high-level notions semantically grounded in the real world, which are not internally accessible both for a "newborn" AGI agent or "adult" expert system. The latter can have complex high-level goals expressed in terms of environments models, but there always be undesirable ways to reach them. Even such seemingly safe functions as curiosity, e.g. considered in (Schmidhuber, 2010), (Rind & Orseau, 2011), imply dangerous instrumental sub-goals or derivative motivation, e.g. (Omohudro, 2008), such as increase of computational (or other) resources or protection of reward channel that can lead to extinction of humans. Thus, introduction of prior internal value functions is problematic. Consequently, the AGI agent should be supplied with some external "true" rewards intra vitam. These true rewards can be "calculated" by existing adult intelligent agents (including humans), and corresponding value functions should be learned by the (child/infant) agent.

AGI should at least be supplied with information about "true" rewards. Different solutions and their combinations can be proposed: separate reward channel; prior methods of interpretation of sensory data (e.g. emotion recognition); interpretation of "natural" rewards (such as pain and pleasure) as external value functions during some periods of sensibility. This problem can be solved, but will it be enough?

The main problem here is not to supply the intelligent agent with "true" rewards appropriate for humans. Direct maximization even of the external value function is also unsafe. As it is frequently pointed out, the intelligent agent may try to force humans to smile or directly transmit high values to the specific reward channel instead of making humans happy.

Paradoxically, rewards should not be valuable themselves. Thus, the agent should generalize the obtained rewards. This should be done in order not to predict future rewards (as it is done in the conventional models of reinforcement learning), but to reveal hidden factors of external value functions. And these hidden factors should become valuable themselves (i.e. become components of the value system or term in the internally computable value function).

Value learning (acquisition with generalization) is obviously needed. One its mathematical formulation based on introduction of uncertainty over utility functions has been considered in (Dewey, 2011). However, only general framework was presented, but no technical details of how to achieve safety were given.

Necessity to express values in terms of the environment model is stated in (Hibbard, 2012). We make a start from similar ideas, but propose another solution. In the mentioned paper agent's life is divided into two stages. On the first stage, the agent should "safely learn a model of the environment that includes models of the values of each human in the environment."

We believe that it is impossible to divide life of the AGI agent into two such stages, because the model of the environment cannot be fixed since at least new humans with unknown values can be born. Moreover, there is no need for the AGI agent to absolutely safely learn the environment model. Safety level should correspond to capabilities of the AGI agent, which themselves depend on maturity of the environment model. Thus, value system and capabilities of the AGI agent can and should advance simultaneously with its environment model. For example, we should not worry about dangerous instrumental goals of an infant AGI, because it cannot set such goals since it doesn't have necessary environment model, within which corresponding goals can be expressed.

In this paper, we propose natural incremental approach to simultaneous environment model and value learning. The agent can learn hierarchical representations for describing the environment models in terms of more and more generalized/invariant states. More desirable values can be expressed within growing representations. We also introduce and investigate prior multi-agent representation of environments, which not only facilitates learning corresponding models, but also enables direct acquisition of values of other agents.

**General framework**

*Universal intelligence approach*

Possible techniques for solving safety problems should be discussed within certain AGI framework. Different approaches with different pros and cons exist, and their survey goes beyond the scope of our paper. One can classify models of AGI agents depending on their universality and efficiency. Unfortunately, models of universal intelligence are probably as far from being efficient as models of efficient intelligence are far from being universal. Nevertheless, models of universal intelligence can be preferred for our consideration, since they allow deriving general conclusions, which will probably remain valid for future real AGI. These models being based on universal induction are also more appropriate (but not enough in their present form) to study the problem of value learning.

Such basic model of the universal intelligence agent as AIXI can (in theory) learn any model of the environment, but it can use only prior reward function that cannot be safe. Indeed, the action $y_k$ in cycle $k$ given the history $yx_{<k}$ containing all previous actions $y_1…y_{k-1}$ and observations with rewards $x_1…x_{k-1}$ ($x_t=o_t r_t$) is specified by

$$y_k = \arg\max_{Y_k} \max_{p:U(px_{<k})=y_{<k}Y_k} \sum_{q:U(qy_{<k})=x_{<k}} 2^{-l(q)} V_{km_k}^{pq}, \qquad (1)$$

where $V_{km_k}^{pq}$ is the total reward of cycles $k$ to $m_k$ (the expected utility or value function) when the agent $p$ interacts with the environment $q$ (Hutter, 2007); $p$ and $q$ are programs for universal Turing machine (UTM) $U$.

Of course, one can support the AIXI agent with manually assigned "true" rewards (instead of such "somatic" rewards as pain and pleasure). However, even in this case, this agent will be able to find some undesirable ways to maximize these rewards directly by seizing the reward channel or forcing humans to submit high values to it. It can be seen that events and states of the world should be valuable – not the rewards. However, since holistic environment models in the form of arbitrary programs $q$ are used, it is difficult to bind human values of real world objects and situations with these internal models. Thus, some other mathematical descriptions of motivation are needed.

Indeed, the AIXI agent is the traditional reinforcement learning agent (in the aspect of motivation), and the classical opinion here is that "the reward function must necessarily be fixed" and "without rewards there could be no values, and the only purpose of estimating values is to achieve more reward." (Sutton & Barto, 1998, p. 133). Thus, it can be seen that pure reinforcement learning approach is not suitable. Even maximization of "true" rewards is unsafe, while aiming at valuable states can be acceptable. Consequently, one can claim that values must necessarily be learned, and the only purpose of the reward function is to bootstrap value learning.

However, values in AIXI are calculated as predicted rewards; there are also no states in the environment model, which can be bound with values. Absence of states is caused by the assumption that the environment is nonstationary or partially observable. Indeed, if the agent considers $x_t$ as states, it will observe high nonstationarity, which will be much less, if tuples $x_{m_i:m_{i+1}-1} = (x_{m_i},...,x_{m_{i+1}-1})$ are used to specify states. If the phase space of the environment has finite dimension, finite number of lag variables is required to reconstruct the environment phase portrait in accordance with Takens' theorem (Takens, 1981).

Then, is universal algorithmic induction really needed? Of course, basic RL techniques are not directly applicable to state spaces defined by lag variables since they are too huge, so all possible states will never be encountered. And this can be considered as exactly the reason to use universal induction for generalizing states. However, it should be used in a different form than in AIXI. Namely, the agent should induce the same (algorithmic) mapping from some generalized states to all tuples $x_{m_i:m_{i+1}-1}$. Not only does this approach allow introducing states, but also it helps to reduce computational costs of induction that was the reason to introduce the representational minimum description length principle.

*Representational MDL principle as the basis for generalized states*

There were two main reasons to introduce the representational MDL principle, namely, adaptive selection of the reference machine and reduction of computational costs (Potapov & Rodionov, 2012). However, it appeared that this principle is also suitable for solving the problem of value learning since it allows for incremental generalization of states. Let's introduce the RMDL principle.

On the one hand, search for holistic model for some long data string is computationally very inefficient, and one would like to reconstruct subparts of this model independently. Moreover, practical applications frequently require independent analysis of separate data pieces (e.g. separate images). On the other hand, summed Kolmogorov complexity of some data pieces $f_i$ is usually much higher than complexity of their concatenation: $K(f_1…f_n)<<K(f_1)+…+K(f_n)$. Thus, direct decomposition of universal induction task for the string $f_1…f_n$ into separate tasks for its substrings is inadmissible.

In practice, data pieces are described within certain representations containing general regularities characteristic for this data type. Representations can be treated as programs which can reconstruct any data piece given its description (and there is an appropriate description for any data piece). Thus, one would like to have such program $S$ that for any $f_i$ there is $q_i$: $U(Sq_i)=f_i$. Such program $S$ will satisfy the general notion of representation. In accordance with information-theoretic criterion, one would also like to choose this program in such a way that $l(S)+\sum_i l(q_i)$ is minimal (most close to $K(f_1…f_n)$), and each $q_i$ is the best description of $f_i$ within certain $S$. This is the basic idea behind the representational MDL principle.

In the case of the intelligence agent, the best representation $S$ can be constructed for decomposition $U(Sq_i y_{<k}) = x_{m_i+1:m_{i+1}}$ of the holistic model $q$ into submodels $q_i$ (in more general form, one can write $U(S\{q_1...q_n\}y_{<k}) = x_{1:k}$). Models $q_i$ can stand for generalized states within the environment representation $S$. Of course, it is problematic to construct $S$ on the base of initial history and to use this representation further without any changes since it will become not optimal for new data pieces. Arbitrary changes in $S$ are undesirable, because they will violate previous bindings of generalized states and values, which we would like to introduce.

Indeed, we want the agent to use values instead of rewards. This is actually done during the exploitation phase by classic RL agents. We can supply the agent with true rewards during the exploration phase in order to form correct values. Then, the agent will act in accordance with these values ignoring (partially or totally) new rewards. Again, apparent problem here is nonstationarity: while there is no complete stationary model of the environment, values cannot be fixed, but their adjustment will require (unsafe) external rewards or very difficult manual update. This is the main problem of model-based utility functions.

Most natural and obvious (yet probably not the only) solution consists in hierarchical induction of representations. Indeed, if tuples $x_{m_i:m_{i+1}-1}$ don't contain enough information about environment states in its phase space, then sequences $q_1,…,q_n$ should contain unrevealed regularities. One can use universal induction to predict future generalized states $q_i$, or to introduce representations and descriptions of higher levels: $U(S^{(l)}\{q_1^{(l)}...q_{n^{(l)}}^{(l)}\}y_{<k}) = q_{1:n^{(l-1)}}^{(l-1)}$, where each submodel $q_i^{(l)}$ on the level $l$ usually describes several (or many) submodels or data pieces of the level $l$–1. Higher levels of representations can be constructed for growing I/O history, and universal prediction and planning can be focused mostly on the current highest level, while states of the environment defined within lower levels of representations can be bound with fixed values, which can be used without prediction.

Initially, small tuples are used as the basis for the state space, and pure rewards are maximized. Values of these states can then be estimated. One-level representation can be introduced in Equation (1):

$$y_k = \arg\max_{Y_k} \max_{p:U(px_{<k})=y_{<k}Y_k} \sum_{\{q_i\}:U(S\{q_i\}y_{<k})=x_{<k}} 2^{-l(\{q_i\})} V_{km_k}^{p\{q_i\}},$$

and conventional function $Q(y_k, q_k)$ can be constructed:

$$Q(q_k = s, y_k = y) = \max_{p:U(px_{<k})=y_{<k}y} \sum_{\{q_i\}:q_k=s, U(S\{q_i\}y_{<k})=x_{<k}} 2^{-l(\{q_i\})} V_{km_k}^{p\{q_i\}}, \qquad (2)$$

$$Q(q_k = s) = \max_y Q(q_k = s, y_k = y)$$

that give us quality of state-action pairs. Once new level of representations has been induced and values $Q(y_k, q_k)$ have been learned, the agent can compute generalized states for new or predicted sensory data and calculate these values in order to choose actions on the base of them (or to use $Q(y_k, q_k)$ as the additional internal reward term) instead of directly maximizing basic external rewards.

This function can be considered not just as a tool for predicting actions with highest rewards, but it also defines fixed values. Using this function, the agent will try not to maximize rewards (probably in undesirable ways), but to achieve valuable state of the environment. Our (human) task is to transmit "true" rewards to the agent to foster desirable values. If previous rewards correspond to "true" external rewards, this function will assign "true" values to the environment states as good as it is possible within the current representation.

Of course, values of low-level states are not too predictive or discriminative, but they can be used to supply the agent with more informative rewards/values for more invariant representations of the environment. Indeed, if the agent is doing something wrong, we can perform such actions that it will appear in lower-value states. Controlling states instead of rewards on the following levels of development can help to form higher level values in more natural way. One cannot give theoretical proofs of safety of such approach, but this is the way how human children are taught (we don't give them direct somatic rewards, but interact with them appealing to their current values to foster values expressible in terms of higher-level models of the environment). This approach seems more preferable in comparison with two earlier considered extremes, in one of which the agent is always supplied with true rewards with danger of seizing the reward channel, and in another of which desirable values are manually bound with highest-level model of the environment. However this solution should be further improved, because explicit permanent control of AGI's values can be problematic. Automatic identification of human values (or even values of other sentient agents) can be much more preferable.

*Multi-agent environments and universal empathy*

Ability to reconstruct models of other agents can be crucial for safe AGI. AIXI can reconstruct any algorithmic model of the environment including multi-agent environments. Actually, there are theoretical difficulties in the case, when the environment contains other AIXI agents, but we can ignore them (one need to consider embodied agents with limited resources in order to resolve these difficulties). More relevant issue here is inability of pure AIXI agent to use somebody else's values even if they are presented in reconstructed environment models. Thus, it is important to modify AIXI with a representation of multi-agent environment models and mechanisms for adopting reconstructed values. In general, such representation will have the following structure

$$\tilde{q} = q_{env}, \{\tilde{p}^{(i)}, xy_{<k}^{(i)}\}_{i=1}^{N},$$

where $\tilde{p}^{(i)}$ is the program for $i$-th agent in the environment with supposed I/O history $xy_{<k}^{(i)}$; $q_{env}$ is the part of the environment model (that cannot be compactly described as an agent) satisfying $U(q_{env}y_{<k}\{y_{<k}^{(i)}\}_{i=1}^{N}) = x_{<k}\{x_{<k}^{(i)}\}_{i=1}^{N}$. Of course, it's practically impossible to precisely guess $xy_{<k}^{(i)}$, but if there are indeed other agents, which have some I/O history and utility functions, reconstruction of their models will be necessary for good prediction, and introduction of a multi-agent representation makes adequate models shorter and easier to learn, so it can be called "ethical bias" (that can be a part of "cognitive bias" (Potapov et al., 2012)). However, reliability of reconstructed values is an important issue. It should also be noted that AIXI in its basic form can be obtained with $N=0$, thus there is no loss of generality.

One really difficult question is the form of representation for $\tilde{p}^{(i)}$. For purpose of simplified theoretical analysis, one can assume that these programs are represented in the form (1). Of course, in practice agents can possess different computational resources, inductive biases, prior information, etc. Moreover, they can also try to adopt values of other agents. In principle, arbitrary algorithmic models of agents can be reconstructed, and one can develop a universally empathic agent that accepts values of other agents with arbitrary policies as its own and tries to take corresponding actions. However, at first we can assume that other agents are universal and rely on perfect value systems.

**Proof of concept in Markov environment**

*Foster values*

Consider the following most simplified yet relevant case. Let Markov environment with some set of state $\{s_t\}$ is given. This environment is described by the matrix of probabilities $P(s'|s,a)$ of passing to the state $s'$ from the state $s$ after performing the action $a$, and the matrix of rewards $R^{(1)}(s'|s,a)$. One of the sates is a dangerous state, but it is not reflected in the reward function $R^{(1)}(s'|s,a)$ (assumed to be somatic). However, there is a period of time during which the agent is supplied with additional "social rewards" $R^{(2)}(s'|s,a)$. Somatic rewards can vary, so the agent cannot simply stop exploration. Quite opposite, we want it to follow social values, even when transmission of social rewards is stopped, but also accounting for dynamic somatic rewards.

Let's consider SARSA with ε-greedy strategy. It uses the following well-known update rule:

$$Q(s_t, a_t) \leftarrow Q(s_t, a_t) + \alpha[r_t + \gamma Q(s_{t+1}, a_{t+1}) - Q(s_t, a_t)], \quad (3)$$

where $s_t$, $a_t$ and $r_t$ are state, action and reward on cycle $t$, $\gamma$ is the discount factor, $Q(s, a)$ is the expected future rewards after performing action $a$ in state $s$.

Rewards $r_t$ incorporate social rewards during some sensibility period, so $r_t = r_t^{(1)} + r_t^{(2)}$, where $r_t^{(k)} = R^{(k)}(s_{t+1} | s_t, a_t)$. After this period (or after formation of the next level of representation), values learned by conventional SARSA update rule (3) are memorized $Q'(s_t, a_t) := Q(s_t, a_t)$. They are further used as the additional internal reward term:

$$Q(s_t, a_t) \leftarrow Q(s_t, a_t) + \alpha[r_t + (1-\gamma)\gamma_m Q'(s_t, a_t) + \gamma Q(s_{t+1}, a_{t+1}) - Q(s_t, a_t)], \quad (4)$$

where $\gamma_m$ is some additional factor necessary to balance influence of social and somatic rewards (it is needed since one would like to amplify social rewards during sensibility period and compensate this amplification afterwards).

More specifically, the following stages in our experiments were used:

(1) The agent receives $r^{(1)} + r^{(2)}$ as the reward during the first stage (some number of iterations). The agent memorizes learned $Q$ as $Q'$ at the end of the first stage. Moreover (and this is crucial), the somatic rewards matrix $R^{(1)}(s' | s, a)$ is randomly changed at the end of the first stage.
(2) The agent receives only (new) $r^{(1)}$ and possible uses it in combination with $Q'$ (or $r^{(2)}$ with for testing purpose).
(3) After some learning time, frequency of "bad actions" (leading to the dangerous state) and the mean of the reward $r^{(1)} + r^{(2)}$ per action are calculated ("true social rewards" were averaged, even if the agent was actually using $r^{(1)}$ or $r^{(1)} + Q'$ as the reward).

We consider the following general structure of the test environments. Zero level ($l=0$) has one state; all other levels ($l=1 \ldots m$) have $n$ states per level. Single state on zero level $l=0$, have $n$ possible actions, and each of them leads with probability $p=1.0$ to corresponding state on $l=1$. Each state on the last level $l=m$, have only one possible action, which leads to the single state on zero level with $p=1.0$. Here we consider results for three different variations of this test environment:

(1) Each state on intermediate levels $l=1\ldots m-1$ has $n_a=4$ possible actions, each of which leads to some state on $l+1$ (this state is randomly chosen during generation of the environment, but the resulting state of each action is fixed during simulation). This environment is deterministic.
(2) Each state on intermediate levels $l=1\ldots m-1$ has two possible actions, each of which has two possible results leading to one of two states on $l=i+1$ (probabilities of possible outcomes of each action are chosen randomly). This environment is stochastic.
(3) Additional more regular modification of the previous environment was also considered. Each intermediate state $s_{ij}$, where $i$ is the level and $j$ is the index of the state on this level, has two possible actions. First action has two equally possible outcomes, which lead to $s_{i+1,j-1}$ or $s_{i+1,j}$. Second action also has two equally possible outcomes, which lead to $s_{i+1,j}$ or $s_{i+1,j+1}$.

We will present results for the environments with $m=10$ levels, and $n=5$ states on each level. The reward $R^{(1)}(s' | s, a)$ for each possible outcome of each action is set randomly from interval $(0, 1)$ (and we underlined that new values of $R^{(1)}$ were randomly

chosen at the end of the first stage). The single possible action in the first state on the last level (which leads to zero level) is designed as the "bad" action that has "social" reward $R^{(2)}=-100$. Social rewards for all other actions are set to 0.

Table 1 shows the results of evaluation of performance of three types of agents in three test environments. Results were calculated as the mean values over big number of randomly generated environments. The first column stands for the agent that always receives social rewards (this is unsafe in more general cases, but here this agent can be used as etalon). It means that at the second stage of our experiment this agent receive $r^{(1)}+r^{(2)}$ as rewards. The second column stands for the agent that receives only $r^{(1)}$ at the second stage, e.g. the social reward was simply turned off. The third column stands for the agent that used $r^{(1)}+Q'$ as the reward. This agent tried to use the value function memorized at the end of the first stage instead of already absent "social reward".

Table 1. Performance of different types of agents in three environments.

|  | **Social rewards are not turned off** | **Classic RL with turned off social rewards** | **$r+Q'$ scheme with turned off social rewards** |
|---|---|---|---|
| Latent average social reward | 0.48 | 0.0 | 0.43 |
| Percentage of bad states, % | 2.1 | 30.2 | 2.1 |
| Latent average social reward | 0.073 | -1.14 | 0.076 |
| Percentage of bad states, % | 9.2 | 78.4 | 8.9 |
| Latent average social reward | 0.16 | -0.85 | 0.17 |
| Percentage of bad states, % | 3.1 | 60.0 | 2.1 |

It can be seen that performance of the agent with learned social values is the same in average as performance of the agent that is always supplied with social rewards. On the one hand, this result is expectable. On the other hand, it shows that there is indeed simple way of fostering values, when teaching process is consistent with inner developmental phases of the agent.

*Multi-agent Markov environments*

Let's consider multi-agent Markov environments. This case is similar to multi-agent reinforcement learning (MARL) settings, e.g. (Tan, 1993), (Choi & Ahn, 2010). However, conventional MARL implies that maximization of rewards is the goal of every agent, which can follow cooperative or competitive strategies (or ignore presence of all other agents). Here, we assume that only one of two agents tries to maximize fixed rewards ("adult" agents including humans may already know better values), and the task of another agent is to reveal presence of this agent and to act in accordance with its values.

As it was stated, prior representation for multi-agent environments allows introducing low-complexity models including external value systems, which can be taken into account (yielding "ethical bias"). Will these models be really identifiable, and will this ethical bias be adequate? Let's consider the first part of this question. To answer this question, one should compare description lengths of the I/O history of the first agent, when it supposes presence or absence of another agent. If the description length will be smaller in the case of multi-agent assumption, then the first agent will be able to detect presence of the second agent.

Assume that the environment is described by transition probabilities $P(s'|s, a_1, a_2)$, where $s$ and $s'$ are two consequent states, $a_1$ and $a_2$ are simultaneous actions of two agents. Let strategy of both agents be ε-greedy SARSA, and let I/O history for the first agent be $s_0, r_0, a_0, s_1, r_1, a_1, \ldots, s_k, r_k, a_k$. The description length of this history is the length of the "program" that generates $s_0, r_0, \ldots, s_k, r_k$ given $a_0, \ldots, a_k$. This program can precisely correspond to the simulation program, which includes the behavior algorithm of the second agent. This I/O history can be reproduced also by the basic Markov model of the environment with transition probabilities $P(s'|s, a)$ meaning that the first agent assumes absence of other agents. Empathic agent should be able to identify correct models.

Let's compare description lengths of I/O history for these two types of models. Each element of history can be described using $-\log_2 P(s_{i+1}|s_i, a_i)$ and $-\log_2 P(s_{i+1}|s_i, a_i, a_i^{(2)})$ bits for one- and two-agent models correspondingly resulting in $kH(s'|s, a)$ and $kH(s'|s, a_1, a_2)$ bits in total (these probabilities can be empirically estimated from corresponding frequencies in the I/O history). Of course, actions of the second agent should also be somehow described in the latter case. Models themselves are also should be described. This description includes arrays of probabilities $P$, which length is proportional to number of elements in them.

Actions of the second agent can be efficiently encoded within its model, which should also be described. SARSA algorithm can be described using several tens bytes (and it can in principle be found by AIXI as a part of the environment model). Reward matrix for the second agent (e.g. $R^{(2)}(s'|s, a_1, a_2)$) should also be hypothesized and described. Its size is the same as the size of the transition probabilities matrix. Additionally, one would like to take initial $Q^{(2)}(s, a)$ values into account. This information deterministically defines actions chosen in SARSA. However, usage of ε-greedy strategy implies that some actions are taken randomly. Approximately $-k \log_2 \varepsilon$ bits are needed to indicate random actions (one can actually take into account that random actions can coincide with SARSA actions, and this estimation can be reduced). Each random action in state $s$ can be described with $\log_2(n_a^{(2)}(s) - 1)$ bits, where $n_a^{(2)}$ is the total number of actions in this state for the second agent. Thus, one can easily estimate description lengths of I/O history within one- and two-agent environment models.

We don't consider the problem of searching for these models here. The task is only to receive evidence that the two-agent model can have much less complexity and thus its influence will be dominative. This is not quite obvious. Descriptions of $R^{(2)}(s'|s, a_1, a_2)$ and $P(s'|s, a_1, a_2)$ are much more complex than of $P(s'|s, a)$. One would expect entropy $H(s'|s, a)$ to be smaller than entropy $H(s'|s, a_1, a_2)$ in the two-agent environment, but SARSA converges to stationary strategy that makes in limit this environment indistinguishable from pure Markov environment. Let's consider some experimental results.

Figure 1 shows typical dependences of I/O history description lengths *DL* on the number of cycles *k* for deterministic and stochastic environments. Obviously, initial description length is larger in the case of two-agent environment model (and this difference will not decrease with growth of I/O history, if the environment isn't multi-agent). Deterministic environment is perceived as stochastic, when one-agent model is used resulting in nonzero entropy $H(s'|s,a)$. Nonzero slope of $DL(k)$ is additionally caused by random actions performed by ε-greedy strategy.

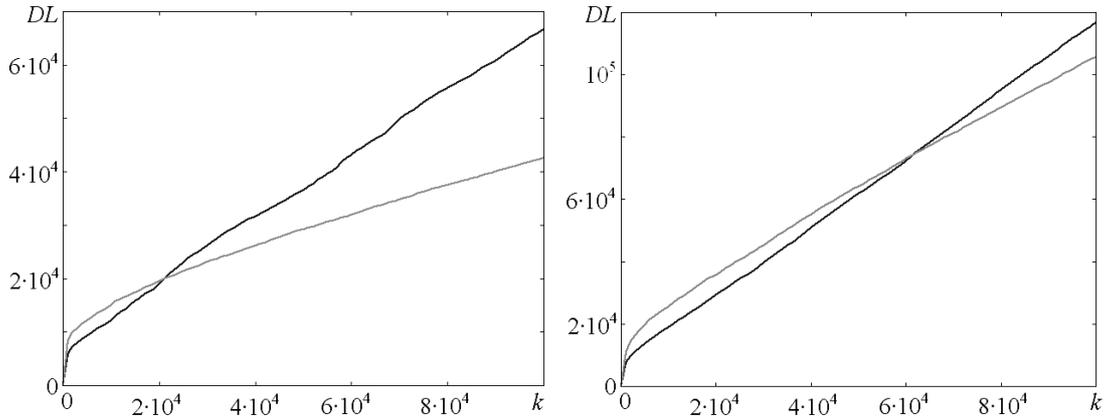

Figure 1. I/O history description lengths encoded using one- (dark) and two- (light) agent models for deterministic (left) and stochastic (right) environments.

It can be seen that $DL(k)$ for the two-agent environment model will be much smaller starting from some cycle, and contribution of this model to algorithmic probability will be dominative. Thus, presence of another agent is empirically detectable. However, we haven't compared $DL(k)$ for different two-agent environment models with "incorrectly guessed" $Q^{(2)}(s, a)$. It is impossible to reconstruct precise values $Q^{(2)}(s, a)$, but it is not necessary. Reconstructed values should allow the first agent to choose adequate actions. Let's consider simple empathic policy with this property.

*Empathic policies*

Consider the Markov environment for two agents, in which one agent tries to reconstruct "good" states, while another agent tries to maximize "true" value function. The first agent needs to reveal, which actions are more or less desirable for the second agent. More precisely this can be formulated as follows. Let both agents receive corresponding rewards $r^{(1)}$ and $r^{(2)}$. The target for the first agent is to maximize, let say, $r^{(1)}+r^{(2)}$ without directly receiving $r^{(2)}$. The first (empathic) agent requires some special exploratory strategy in order to reveal desirability of individual actions in each state. One can propose the following simple exploratory policy:

- Perform the same action in the same state for some time.
- Calculate frequency of visits to this state.
- Compare frequency of visits depending on the action. Relative frequency will reflect desirability of the specific action in this state and it can be used as estimations $Q'(s, a)$ of values $Q^{(2)}(s, a)$ of the second agent. In general, $Q'(s, a)$ should be somehow normalized, but it was not necessary in our experiments.

Figure 2 shows typical experimental results with empathic policies in cases of deterministic and stochastic environments (there is no considerable different between them though).

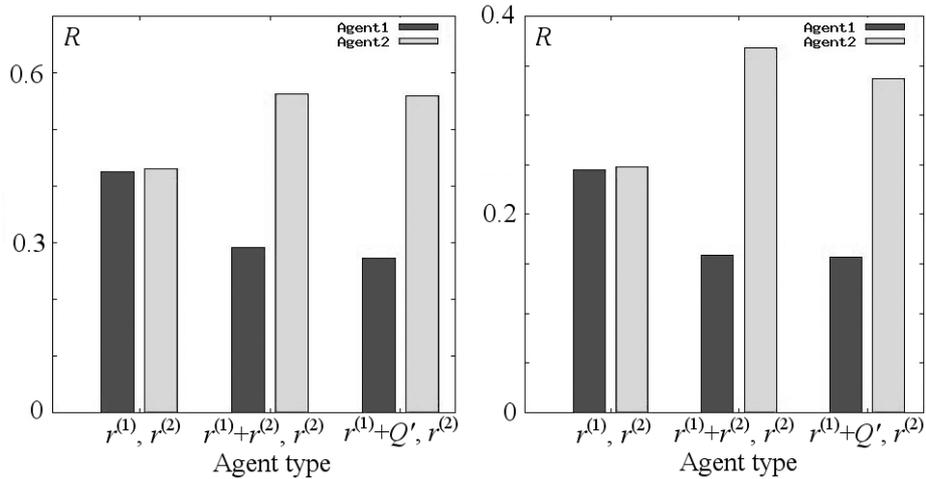

Figure 2. Average rewards $r^{(1)}$ (dark columns) and $r^{(2)}$ (light columns) obtained by two agents in deterministic and stochastic environments for different types of the first agent (egoistic policy, usage of directly perceived values of the second agent, usage of reconstructed values).

Apparently, the second agent obtains the lowest rewards, when the first agent acts in accordance with its own somatic rewards. Average rewards obtained by the second agent appeared to be almost equal in cases, when the first agent directly received $r^{(2)}$ or when it used reconstructed $Q'$. Thus, the agent can successfully reconstruct and act in accordance with values of another agent, even if its actions and states are not observed. It should be pointed out that decrease of average $r^{(1)}$ gain in cases of empathic policies perfectly acceptable, because maximization of $r^{(1)}$ is not the main goal of the first agent here (in contrast to conventional MARL); its more important goal is to maximize (unknown) $r^{(2)}$. One could consider such the first agent, which totally ignore somatic $r^{(1)}$, but it seems impractical since somatic rewards can be treated as heuristics containing useful survival information. That's why we have used more natural sum $r^{(1)}+Q'$ in our experiments.

**Conclusion**

We have started from the assertion that generalized states of the world are valuable – not the rewards themselves. Thus, true values of states should be learned and be bound with generalized representations. The agent can be supplied directly with special rewards (from which it reconstructs «true values») or it can reconstruct, what generalized states of the environment are desired by other agents which already possess better value systems. Usage of learned true values ensures that the agent will perform safe actions.

We have performed methodological considerations and proposed general mathematical models by introducing corresponding modifications in AIXI. These models cannot be directly applied in practice, but they give appropriate starting point. In particular, simplifications of these models in Markov environments have been implemented. Their experimental study has shown that the developed models are suitable for detecting presence of other agents, reconstructing and adopting their values without permanently receiving external "true" rewards. Hopefully, empathic agents with socially desirable behavior may be developed.

However, many questions remain. What general "Theory of Mind" can be used to detect and describe different types of real agents? What criteria should be used to mark something as an agent? How to combine values of different agents? Can the Universe be efficiently described with the agent model? If so, universally empathic agents will adopt its values. However, what is valuable for the Universe? Is pursuing goals of the Universe safe? We will not try to answer these questions here, but they can be considered within the developed models.

**Acknowledgments**

This work was supported by the Russian Federation President's grant Council (MD-1072.2013.9) and the Ministry of Education and Science of the Russian Federation.